\documentclass[10pt,twocolumn,letterpaper]{article}
\usepackage{cvpr}
\usepackage{times}
\usepackage{epsfig}
\usepackage{graphicx}
\usepackage{amsmath}
\usepackage{amssymb}
\usepackage{subfigure}
\DeclareMathOperator*{\argmax}{arg\,max}

\usepackage[pagebackref=true,breaklinks=true,letterpaper=true,colorlinks,bookmarks=false]{hyperref}

\hypersetup{ 
       colorlinks=true,
   linkcolor=black, 
       citecolor=black,
       filecolor=black,
       menucolor=black,
       pagecolor=black,
       urlcolor=black,
       breaklinks=true
}

\cvprfinalcopy % *** Uncomment this line for the final submission

\def\cvprPaperID{****} % *** Enter the CVPR Paper ID here

\begin{document}
%%%%%%%%% TITLE & AUTHOR
\title{SpotTune: Transfer Learning through Adaptive Fine-tuning \vspace{-0.1in}}

\author{Yunhui Guo\thanks{
This work was done when Yunhui Guo was an intern at IBM Research. 
\textsuperscript{$\dag$}Abhishek Kumar is now with Google Brain. The work was done when he was at IBM Research.}~$^{1,2}$,
Honghui Shi$^{1}$, 
Abhishek Kumar$^{\dag 1}$,
Kristen Grauman$^{3}$,
Tajana Rosing$^{2}$,
Rogerio Feris$^{1}$ \vspace{0.05in} \\ 
{\normalsize $^{1}$IBM Research \& MIT-IBM Watson AI Lab},
{\normalsize $^{2}$University of California, San Diego},
{\normalsize $^{3}$The University of Texas at Austin}
}

\maketitle
%\thispagestyle{empty}

%\vspace{-0.5mm}
%%%%%%%%% ABSTRACT
\begin{abstract}
Transfer learning, which allows a source task to affect the inductive bias of the target task, is widely used in computer vision. 
The typical way of conducting transfer learning with deep neural networks is to fine-tune a  model pre-trained on the source task using data from the target task. In this paper, we propose an adaptive fine-tuning approach, called SpotTune, which finds the optimal fine-tuning strategy {\em per instance} for the target data. In SpotTune, given an image from the target task, a policy network is used to make routing decisions on whether to pass the image through the fine-tuned layers or the pre-trained layers. 
We conduct extensive experiments to demonstrate the effectiveness of the proposed approach. Our method outperforms the traditional fine-tuning approach on 12 out of 14 standard datasets.
We also compare SpotTune with other state-of-the-art fine-tuning strategies, showing superior performance. On the {\em Visual Decathlon} datasets, our method achieves the highest score across the board without bells and whistles.

\vspace{-0.1in}

\end{abstract}
%%%%%%%%% BODY TEXT
\section{Introduction}

Deep learning has shown remarkable success in many computer vision tasks, but current methods often rely on large amounts of labeled training data \cite{krizhevsky2012imagenet, he2016deep,huang2017densely}. \emph{Transfer learning}, where the goal is to transfer knowledge from a related \emph{source task}, is commonly used to compensate for the lack of sufficient training data in the \emph{target task} \cite{pan2010survey, bengio2012deep}. \emph{Fine-tuning} is arguably the most widely used approach for transfer learning when working with deep learning models. %This method 
It starts with a pre-trained model on the source task and trains it further on the target task. For computer vision tasks, it is a common practice to work with ImageNet pre-trained models for fine-tuning \cite{kornblith2018better}. Compared with training from scratch, fine-tuning a pre-trained convolutional neural network on a target dataset can significantly improve performance, while reducing the target labeled data %requirement 
requirements~\cite{girshick2014rich, yosinski2014transferable,tajbakhsh2016convolutional, kornblith2018better}.

\begin{figure}[!htb]
    \centering
\includegraphics[width=0.48\textwidth]{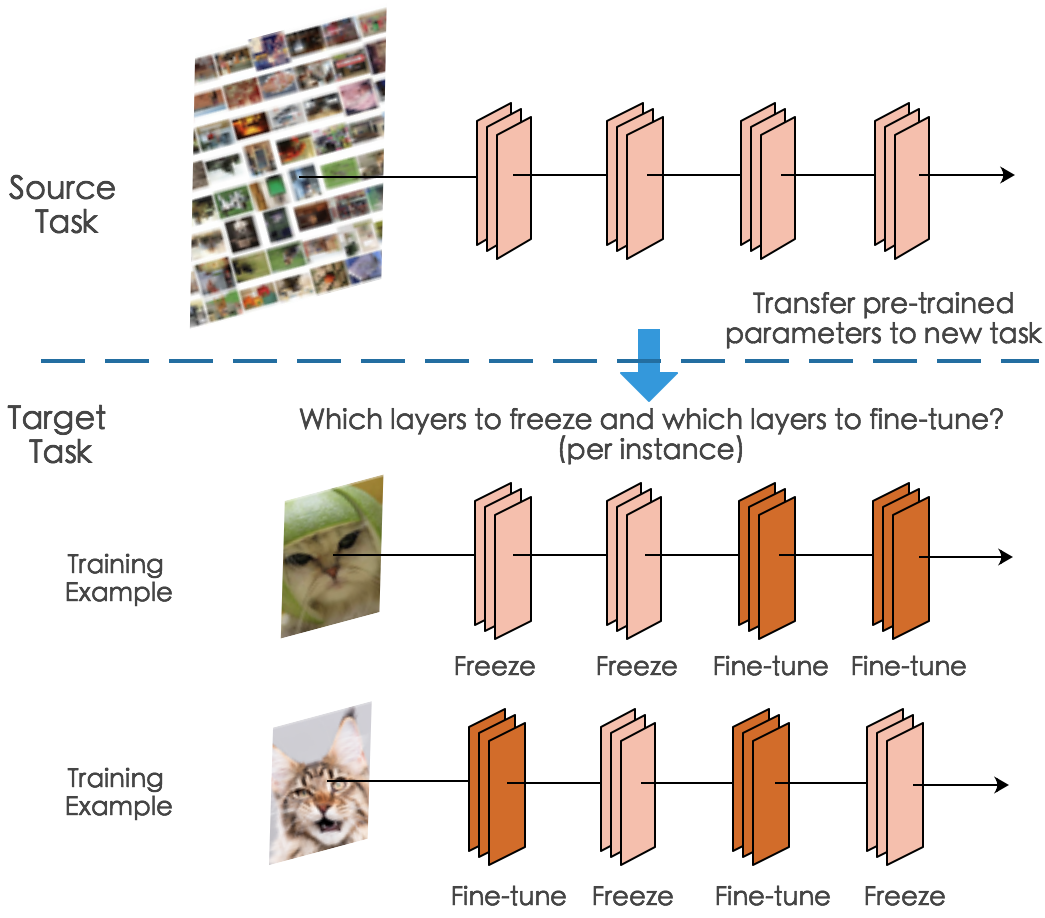}    \caption{Given a deep neural network pre-trained on a source task, we address the question of {\em where to fine-tune} its parameters with examples of the target task. We propose a novel method that decides, per training example, which layers of the pre-trained model should have their parameters fixed, i.e., shared with the source task, and which layers should be fine-tuned to improve the accuracy of the model in the target domain.}
    \label{fig:motivation}
\end{figure}

There are several choices when it comes to realizing the idea of fine-tuning of deep networks in practice. A natural approach is to optimize \emph{all} the parameters of the deep network using the target training data (after initializing them with the parameters of the pre-trained model). However, if the target dataset is small and the number of parameters is huge, fine-tuning the whole network may result in overfitting \cite{yosinski2014transferable}. Alternatively, the last few layers of the deep network can be fine-tuned while freezing the parameters of the remaining initial layers to their pre-trained values \cite{tajbakhsh2016convolutional, azizpour2016factors}. This is driven by a combination of limited training data in the target task and the empirical evidence that initial layers learn low-level features that can be directly shared across various computer vision tasks. However, the number of initial layers to freeze during fine-tuning still remains a manual design choice which can be inefficient to optimize for, especially for networks with hundreds or thousands of layers. Further, it has been empirically observed that current successful multi-path deep architectures such as ResNets \cite{he2016deep} behave like ensembles of shallow networks \cite{veit2016residual}. It is not clear if restricting the fine-tuning to the last contiguous layers is the best option, as the ensemble effect diminishes the assumption that early or middle layers should be shared with common low-level or mid-level features.

Current methods also employ a \emph{global fine-tuning} strategy, \ie, the decision of which parameters to freeze vs fine-tune is taken for all the examples in the target task. 
The assumption is that such a decision is optimal for the entire target data distribution, which may not be true, particularly in the case of insufficient target training data. 
For example, certain classes in the target task might have higher similarity with the source task, and routing these target examples through the source pre-trained parameters (during inference) might be a better choice in terms of accuracy. Ideally, we would like these decisions to be made individually for each layer (\ie, whether to use pre-trained parameters or fine-tuned parameters for that layer), per input example, as illustrated in Figure \ref{fig:motivation}.

In this paper, we propose \emph{SpotTune}, an approach to learn a decision policy for input-dependent fine-tuning. 
The policy is sampled from a discrete distribution parameterized by the output of a lightweight neural network, which decides which layers of a pre-trained model should be fine-tuned or have their parameters frozen, on a per instance basis. As these decision functions are discrete and non-differentiable, we rely on a recent Gumbel Softmax sampling approach \cite{maddison2016concrete,jang2016categorical} to train the policy network. At test time, the policy decides whether the features coming out of a layer go into the next layer with source pre-trained parameters or the fine-tuned parameters.

We summarize our contributions as follows: 
\begin{itemize}
\item We propose an input-dependent fine-tuning approach that automatically determines which layers to fine-tune per target instance. This is in contrast to current fine-tuning methods which are mostly ad-hoc in terms of determining {\em where to fine-tune} in a deep neural network  (\eg, fine-tuning last $k$ layers). 
\item  We also propose a global variant of our approach that constrains all the input examples to fine-tune the same set of $k$ layers which can be distributed anywhere in the network. This variant results in fewer parameters in the final model as the corresponding set of pre-trained layers can be discarded.  
\item We conduct extensive empirical evaluation of the proposed approach, comparing it with several competitive baselines. The proposed approach outperforms standard fine-tuning on 12 out of 14 datasets. Moreover, we show the effectiveness of \emph{SpotTune} compared to other state-of-the-art fine-tuning strategies. On the Visual Decathlon Challenge \cite{rebuffi2017learning}, which is a competitive benchmark for testing the performance of multi-domain learning algorithms with a total of 10 datasets, the proposed approach achieves the highest score compared with the state-of-the-art methods.
\end{itemize}

%------------------------------------------------------------------------
\section{Related Work}
{\bf Transfer Learning.} There is a long history of transfer learning and domain adaptation methods in computer vision \cite{csurkasurvey,pan2010survey}. Early approaches have concentrated on {\em shallow classifiers}, using techniques
such as instance re-weighting \cite{tradaboost, dudik2006correcting}, model adaptation \cite{duan2009domain,yang2007cross}, and feature space alignment \cite{sun2016return,mirrashed2013domain,gong2012geodesic}. In the multi-task setting, knowing which tasks or parameters are shareable is a longstanding challenge~\cite{kang2011learning,kumar2012learning,thrun1998clustering,lu2017fully}.
More recently, transfer learning based on deep neural network classifiers has received significant attention in the community \cite{ganin2016domain, chen2015deep, dlid, co-regularized,ge2017borrowing}. Fine-tuning a pre-trained network model such as ImageNet on a new dataset is the most common strategy for knowledge transfer in the context of deep learning. Methods have been proposed to fine-tune all network parameters \cite{girshick2014rich}, only the parameters of the last few layers \cite{long2015learninglastfew}, or to just use the pre-trained model as a fixed feature extractor with a classifier such as SVM on top \cite{sharif2014cnn}. Kornblith et al. \cite{kornblith2018better} have studied several of these options to address the question of whether better ImageNet models transfer better. Yosinski et al. \cite{yosinski2014transferable} conducted a study on the impact of transferability of features from the bottom, middle, or top of the network with early models, but it is not clear whether their conclusions hold for modern multi-path architectures such as Residual Networks \cite{he2016deep} or DenseNets \cite{huang2017densely}. Yang et al. \cite{yang2018glomo} have recently proposed to learn relational graphs as transferable representations, instead of unary features. More related to our work,  Li et al. \cite{li2018explicit} investigated several regularization schemes that explicitly promote the similarity of the fine-tuned model with the original pre-trained model. Different from all these methods, our proposed approach automatically decides the optimal set of layers to fine-tune in a pre-trained model on a new task. In addition, we make this decision on a {\em per-instance} basis.  

\vspace{0.05in}
{\bf Dynamic Routing.} Our proposed approach is related to {\em conditional computation} methods \cite{bengio2013estimating,liu2017dynamic,figurnov2017spatially}, which
aim to dynamically route information in neural networks with the goal of improving computational efficiency. Bengio et al.~\cite{bengio2015conditional} used sparse activation policies to selectively execute neural network units on a per-example basis. Shazeer et al.~\cite{shazeer2017outrageously} introduced a Sparsely-Gated Mixture-of-Experts layer, where a trainable gating network determines a sparse combination of sub-networks (experts) to use for each example. Wu, Nagarajan et al.~proposed {\em BlockDrop} \cite{wu2018blockdrop}, a method that uses reinforcement learning to dynamically select which layers of a Residual Network to execute, exploiting the fact that ResNets are resilient to layer dropping \cite{veit2016residual}. Veit and Belongie \cite{veit2018convolutional} investigated the same idea using Gumbel Softmax \cite{jang2016categorical} for on-the-fly selection of residual blocks. Our work also %relies on 
explores 
dynamic routing based on the {\em Gumbel trick}. However, unlike previous methods, our goal is to determine the parameters in a neural network that should be frozen or fine-tuned during learning to improve accuracy, instead of dropping layers to improve efficiency.

%------------------------------------------------------------------------
\section{Proposed Approach}

\begin{figure*}
    \centering
\includegraphics[width=1.0\textwidth]{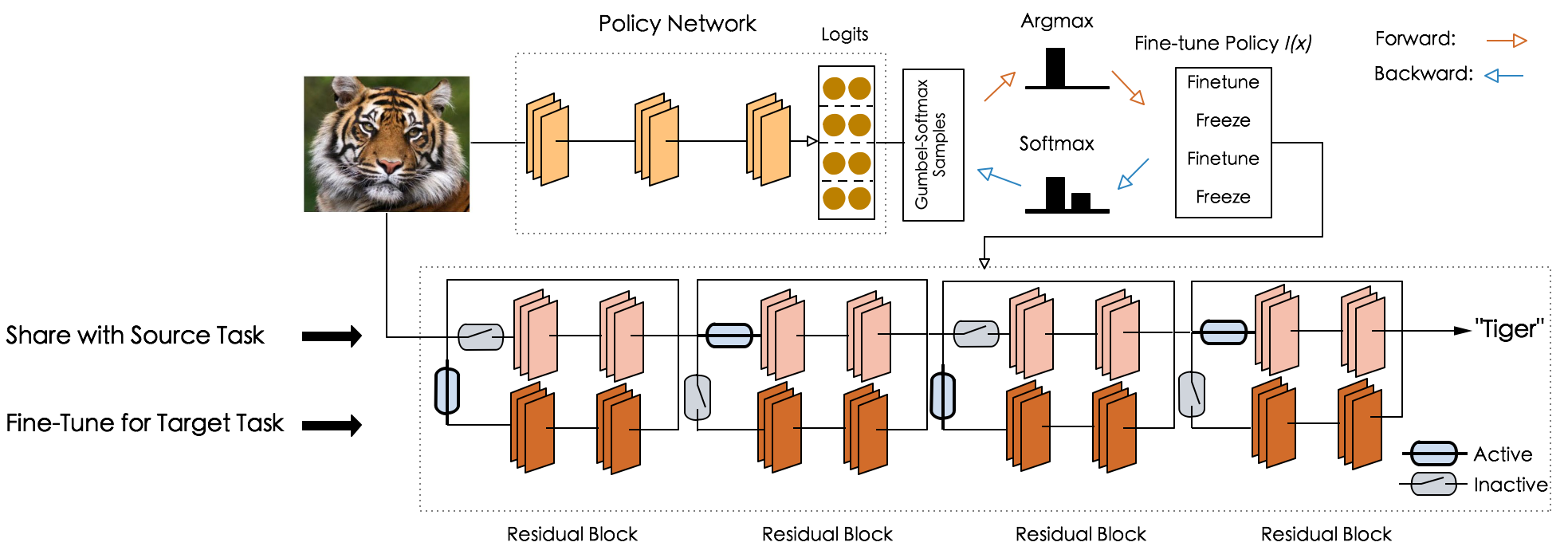}  
    \caption{Illustration of our proposed approach. The policy network is trained to output routing decisions (fine-tune or freeze parameters) for each block in a ResNet pre-trained on the source dataset. During learning, the fine-tune vs. freeze decisions are generated based on a Gumbel Softmax distribution, which allows us to optimize the policy network using backpropagation. At test time, given an input image, the computation is routed so that either the fine-tuned path or the frozen path is activated for each residual block. }
    \label{fig:spottune}
\end{figure*}

Given a pre-trained network model on a source task (e.g., ImageNet pre-trained model), and a set of training examples with associated labels in the target domain, our goal is to create an adaptive fine-tuning strategy that decides, per training example, which layers of the pre-trained model should be fine-tuned (adapted to the target task) and which layers should have their parameters frozen (shared with the source task) during training, in order to improve the accuracy of the model in the target domain. 
To this end, we first present an overview of our approach in Section \ref{sec: spottune}. Then, we show how we learn our adaptive fine-tuning policy using Gumbel Softmax sampling in Section \ref{sec: gumbel}. Finally, in Section \ref{sec:global}, we present a global policy variant of our proposed image-dependent fine-tuning method, which constraints all the images to follow a single fine-tuning policy.

\subsection{SpotTune Overview}
\label{sec: spottune}
Although our approach could be applied to different deep neural network architectures, in the following we focus on a Residual Network model (ResNet) \cite{he2016deep}. Recently, it has been shown that ResNets behave as ensembles of shallow classifiers and are resilient to residual block swapping \cite{veit2016residual}. This is a desirable property for our approach, as later we %will 
show that SpotTune dynamically swaps pre-trained and fine-tuned blocks to improve performance.

Consider the $l$-th residual block in a pre-trained ResNet model:
\begin{equation}
    x_l = F_l(x_{l-1}) + x_{l-1}.
\end{equation}

In order to decide whether or not to fine-tune a residual block during training, we \textit{freeze} the original block %$F_l(x_{l-1})$ 
$F_l$ and create a new \textit{trainable} block $\hat{F_l}$, which is initialized with the parameters of $F_l$. With the additional block $\hat{F_l}$, the output of the $l$-th residual block in SpotTune is computed as below:
\begin{equation}
\begin{aligned}
    &x_l = I_{l}(x)\hat{F}_l(x_{l-1}) + (1-I_{l}(x))F_l(x_{l-1}) + x_{l-1}  \\
\end{aligned}
\end{equation}

\noindent where $I_l(x)$ is a binary random variable, which indicates whether the residual block should be frozen or fine-tuned,  conditioned on the input image. During training, given an input image $x$, the \textit{frozen} block $F_l$ trained on the source task is left unchanged and the replicated block $\hat{F_l}$, which is initialized from $F_l$, can be optimized towards the target dataset. Hence, the given image $x$ can either share the \textit{frozen} block $F_l$, which allows the features computed on the source task to be reused, or fine-tune the block $\hat{F_l}$, which allows $x$ to use the adapted features. $I_l(x)$ is sampled from a discrete distribution with two categories (freeze or fine-tune), which is parameterized by the output of a lightweight policy network. More specifically, if $I_l(x)$ = 0, then the $l$-th frozen block is re-used. Otherwise, if $I_l(x)$ = 1 the $l$-th residual block is fine-tuned by optimizing $\hat{F_l}$.

Figure \ref{fig:spottune} illustrates the architecture of our proposed \textit{SpotTune} method, which allows each training image to have its own fine-tuning policy. During training, the policy network is jointly trained with the target classification task using Gumbel Softmax sampling, as we will describe next. At test time, an input image is first fed into a policy network, whose output is sampled to produce routing decisions on whether to pass the image through the fine-tuned or pre-trained residual blocks. The image is then routed through the corresponding residual blocks to produce the final classification prediction. Note that the effective number of executed residual blocks is the same as the original pre-trained model. The only additional computational cost is incurred by the policy network, which is designed to be lightweight (only a few residual blocks) in comparison to the original pre-trained model.

\subsection{Training with the Gumbel Softmax Policy}
\label{sec: gumbel}
SpotTune makes decisions as to whether or not to freeze or fine-tune each residual block per training example. However, the fact that the policy $I_l(x)$ is discrete makes the network non-differentiable and therefore difficult to be optimized with backpropagation. There are several ways that allow us to ``back-propagate'' through the discrete nodes \cite{bengio2013estimating}. In this paper, we use a recently proposed Gumbel Softmax sampling approach \cite{maddison2016concrete,jang2016categorical} to circumvent this problem. 

The Gumbel-Max trick \cite{maddison2016concrete} is a simple and effective way to draw samples from a categorical distribution parameterized by $\{\alpha_1, \alpha_2, ..., \alpha_z\}$, where $\alpha_i$ are scalars not confined to the simplex, and $z$ is the number of categories. In our work, we consider two categories (freeze or fine-tune), so $z=2$, and for each residual block, $\alpha_1$ and $\alpha_2$ are scalars corresponding to the output of a policy network.  

A random variable $G$ is said to have a standard Gumbel distribution if $G = -\log (- \log(U))$ with $U$ sampled from a uniform distribution, i.e. $U \sim Unif[0,1]$. Based on the Gumbel-Max trick \cite{maddison2016concrete}, we can draw samples from a discrete distribution parameterized by $\alpha_i$ in the following way: we first draw i.i.d samples $G_i, ..., G_z$ from $Gumbel(0,1)$ and then generate the discrete sample as follows:
\begin{equation} 
    X = \argmax_{i} [\log \alpha_i + G_i].
    \label{Eq: forward}
\end{equation}
 
  The $\argmax$ operation in Equation \ref{Eq: forward} is non-differentiable. However, we can use the Gumbel Softmax distribution  \cite{maddison2016concrete,jang2016categorical}, which adopts softmax as a continuous relaxation to $\argmax$. We represent $X$ as a one-hot vector where the index of the non-zero entry of the vector is equal to $X$, and relax the one-hot encoding of $X$ to a $z$-dimensional real-valued vector $Y$ using softmax:
 
 \begin{equation}
     Y_i = \frac{\exp((\log \alpha_i+ G_i)/\tau)} {\sum_{j=1}^{z} \exp((\log \alpha_j + G_j) / \tau)}  \quad \textnormal{for } i = 1,..,z
     \label{Eq: backward}
 \end{equation}
 where $\tau$ is a temperature parameter, which controls the discreteness of the output vector $Y$. When $\tau$ becomes closer to 0, the samples from the Gumbel Softmax distribution become indistinguishable from the discrete distribution (i.e, almost the same as the one-hot vector). 
 
 Sampling our fine-tuning policy $I_l(x)$ from a Gumbel Softmax distribution parameterized by the output of a policy network allows us to backpropagate from the discrete freeze/fine-tune decision samples to the policy network, as the Gumbel Softmax distribution is smooth for $\tau > 0$ and therefore has well-defined gradients with respect to the parameters $\alpha_i$. By using a standard classification loss $l_c$ for the target task, the policy network is jointly trained with the pre-trained model to find the optimal fine-tuning strategy that maximizes the accuracy of the target task.
 
 Similar to \cite{wu2018blockdrop}, we generate all freeze/fine-tune decisions for all residual blocks at once, instead of relying on features of intermediate layers of the pre-trained model to obtain the fine-tuning policy. More specifically, suppose there are $L$ residual blocks in the pre-trained model. The output of the policy network is a two-dimensional matrix $\beta \in \mathbb{R}^{L \times 2}$. Each row of $\beta$ represents the logits of a Gumbel-Softmax Distribution with two categories, i.e, $\beta_{l,0} = \log \alpha_1$ and $\beta_{l,1} = \log \alpha_2$. After obtaining $\beta$, we use the straight-through version of the Gumbel-Softmax estimator \cite{jang2016categorical}. During the forward pass, we sample the fine-tuning policy $I_l(x)$ using Equation \ref{Eq: forward} for the $l$-th residual block. During the backward pass, we approximate the gradient of the discrete samples by computing the gradient of the continuous softmax relaxation in Equation \ref{Eq: backward}. This process is illustrated in Figure \ref{fig:spottune}.
 
\subsection{Compact Global Policy Variant}
\label{sec:global}
In this section, we consider a simple extension of the image-specific fine-tuning policy, which constrains all the images to fine-tune the same $k$ blocks that can be distributed anywhere in the ResNet. This variant reduces both the memory footprint and computational costs, as $k$ can be set to a small number so most blocks are shared with the source task, and at test time the policy network is not needed.

Consider a pre-trained ResNet model with $L$ residual blocks. For the $l$-th block, we can obtain the number of images that use the fine-tuned block and the pre-trained block based on the image-specific policy. We compute the fraction of images in the target dataset that uses the fine-tuned block and denote it as $v_l \in [0, 1]$. In order to constrain our method to fine-tune $k$ blocks, we introduce the following loss:
\begin{equation}
    l_{k} = ((\sum_{l=1}^Lv_l) - k)^2.
\end{equation}

Moveover, in order to achieve a deterministic policy, we add another loss $l_{e}$:
\begin{equation}
    l_{e} = \sum_{l=1}^L -v_l\log v_l.
\end{equation}

The additional loss $l_{e}$ pushes $v_l$ to be exactly 0 or 1, so that a global policy can be obtained for all the images. The final loss is defined below:

\begin{equation}
    l = l_c + \lambda_1 l_{k} + \lambda_2 l_{e},
\end{equation}

\noindent where $l_c$ is the classification loss, $\lambda_1$ is the balance parameter for $l_{k}$, and $\lambda_2$ is the the balance parameter for $l_{e}$. The additional losses push the policy network to learn a global policy for all the images. %Different from 
As opposed to manually selecting $k$ blocks to fine-tune, the global-k variant {\em learns} the $k$ blocks that can achieve the best accuracy on the target dataset. We leave for future work the task of finding the optimal $k$, which could be achieved e.g., by using reinforcement learning with a reward proportional to accuracy and inversely proportional to the number of fine-tuned blocks.

%------------------------------------------------------------------------
\section{Experiments}
\subsection{Experimental Setup}
\noindent \textbf{Datasets and metrics.} 
We compare our SpotTune method with other fine-tuning and regularization techniques on 5 public datasets, including three fine-grained classification benchmarks: CUBS \cite{wah2011caltech}, Stanford Cars \cite{krause20133d} and Flowers \cite{nilsback2008automated}, and two datasets with a large domain mismatch from ImageNet: Sketches \cite{eitz2012humans} and WikiArt \cite{saleh2015large}. The statistics of these datasets are listed in Table \ref{table: datasets}. Performance is measured by classification accuracy on the evaluation set.

We also report results on the datasets of the Visual Decathlon Challenge \cite{rebuffi2017learning}, which aims at evaluating visual recognition algorithms on images from multiple visual domains. There are a total of 10 datasets as part of this challenge: (1) ImageNet, (2) Aircraft, (3) CIFAR-100, (4)  Describable textures, (5) Daimler pedestrian classification, (6) German traffic signs, (7) UCF-101 Dynamic Images, (8) SVHN, (9) Omniglot, and (10) Flowers. The images of the Visual Decathlon datasets are resized isotropically to have a shorter side of 72 pixels, in order to alleviate the computational burden for evaluation.  
Following \cite{rebuffi2017learning}, the performance is measured by a single scalar score $S = \sum_{i=1}^{10}\alpha_i \textnormal{max}\{0, E_i^{\textnormal{max}}-E_i\}^2$, 
where $E_i$ is the test error on domain $D_i$, and $E_i^{\textnormal{max}}$ is the error of a reasonable baseline algorithm. The coefficient $\alpha_{i}$ is  $1000(E_i^{\textnormal{max}})^{-2}$, so a perfect classifier receives score 1000. The maximum score achieved across 10 domains is 10000. Compared with average accuracy across all the 10 domains, the score $S$ is a more reasonable measurement for comparing different algorithms, since it considers the difficulty of different domains, which is not captured by the average accuracy \cite{rebuffi2017learning}.

In total, our experiments comprise 14 datasets, as the Flowers dataset is listed in both sets described above. We note that for the experiments in Table \ref{table:results}, we use the full resolution of the images, while those are resized in the Visual Decathlon experiments to be consistent with other approaches.\begin{center}
\begin{table}[t]
\center
\small
\begin{tabular}{ |c|c|c|c| } 
 \hline
 \textbf{Dataset} &  \textbf{Training} & \textbf{Evaluation} & \textbf{Classes} \\ 
\hline
CUBS & 5,994& 5,794&200 \\
\hline
Stanford Cars & 8,144& 8,041& 196\\
\hline
Flowers & 2,040 & 6,149& 102\\
\hline
Sketch & 16,000& 4,000& 250\\
\hline
WikiArt & 42,129& 10,628& 195\\
\hline
\end{tabular}
\caption{Datasets used to evaluate SpotTune against other fine-tuning baselines.}
\label{table: datasets}
\end{table}
\end{center}
\def\arraystretch{1.1}%  1 is the default, change whatever you need
\begin{table*}[!htb]
	\small
	\begin{center}
		\begin{tabular}{|c| c| c| c| c|c|} 
			\hline
		    Model & CUBS & Stanford Cars &  Flowers & WikiArt & Sketches \\
            \hline
        	 Feature Extractor& 74.07\% & 70.81\% & 85.67\%  & 61.60\%  & 75.50\% \\
        	 \hline
        	Standard Fine-tuning & 81.86\% & 89.74\% & 93.67\%  &  75.60\%& 79.58\% \\
            \hline      
            
            Stochastic Fine-tuning & 81.03\% & 88.94\%  & 92.95\%   & 73.06\% & 78.30\% \\
            \hline      
            
            Fine-tuning last-3& 81.54\% &  88.21\% & 89.03\%  &  72.68 \%  & 77.72\%  \\
            
            \hline      
            Fine-tuning last-2& 80.34\% & 85.36\%  &  91.81\% &70.82\%  & 78.37\% \\
            \hline   
    
            Fine-tuning last-1& 78.68\% & 81.73\%  & 89.99\%  & 68.96\%  & 77.20\% \\
            \hline   
            
            Fine-tuning ResNet-101& 82.13\% & 90.32\% & 94.21\%  & \textbf{76.52\%}  & 78.92\% \\
            \hline
            $L^2$-SP  & 83.69\% & 91.08\%  & 95.21\%   & 75.38\%  & 79.60\%  \\
              \hline   
            SpotTune (running fine-tuned blocks) & 82.36\% & 92.04\%  & 93.49\%  & 67.27\% & 78.88\% \\
             \hline   
            SpotTune (global-k) &  83.48\% & 90.51\% &  \textbf{96.60\%}  &  75.63\% &  80.02\%\\
             \hline   
            SpotTune & \textbf{84.03} \% & \textbf{92.40\%}  & 96.34\%  &  75.77\% &  \textbf{80.20\%} \\
            \hline   
        
		\end{tabular}
	\end{center}
		\caption{Results of \emph{SpotTune} and baselines on CUBS, Stanford Cars, Flowers, WikiArt and Sketches.}
	\label{table:results}
\end{table*}
\vspace{-0.4cm}
\noindent \textbf{Baselines.} We compare SpotTune with the following fine-tuning and regularization techniques:
\begin{itemize} 
  \setlength\itemsep{0.01em}
    \item { \textbf{Standard Fine-tuning}}: This baseline fine-tunes all the parameters of the pre-trained network on the target dataset \cite{girshick2014rich,yosinski2014transferable}. 
    
    \item { \textbf{Feature Extractor}}: We use the pre-trained network as a feature extractor~\cite{sharif2014cnn,donahue2014decaf} and only add the classification layer for each newly added dataset.

    \item { \textbf{Stochastic Fine-tuning}}: We randomly sample 50\% of the blocks of the pre-trained network to fine-tune.
    
    \item { \textbf{Fine-tuning last-k} ($k$ = 1, 2, 3)}: This baseline fine-tunes the last $k$ residual blocks of the pre-trained network on the target dataset \cite{long2015learninglastfew,tajbakhsh2016convolutional, azizpour2016factors}. In our experiments, we consider fine-tuning the last one ($k$ = 1), last two ($k$ = 2) and the last three ($k$ = 3) residual blocks. 
    
    \item { \textbf{Fine-tuning ResNet-101}}: We fine-tune all the parameters of a pre-trained ResNet-101 model on the target dataset. SpotTune uses ResNet-50 instead (for the experiments in Table \ref{table:results}), so this baseline is more computationally expensive and can fine-tune twice as many residual blocks. %But we decided to
    We include it as the total number of parameters during training is similar to SpotTune, so it will verify any advantage is not merely due to our having 2x residual blocks available.
    
    \item { \textbf{$L^2$-SP} \cite{li2018explicit}}: This is a recently proposed state-of-the-art regularization method for fine-tuning. The authors recommend using an $L^2$ penalty to allow the fine-tuned network to have an explicit inductive bias towards the pre-trained model, sharing similar motivation with our approach.
\end{itemize}

Regarding the methods that have reported results on the Visual Decathlon datasets, the most related to our work are models trained from {\em Scratch}, {\em Standard Fine-tuning}, the {\em Feature Extractor} baseline as described above, and {\em Learning without Forgetting (LwF)} \cite{li2017learning}, which is a recently proposed technique that encourages the fine-tuned network to retain the performance on ImageNet or previous tasks, while learning consecutive tasks. Other methods include {\em Piggyback} \cite{mallya2018piggyback}, {\em Residual Adapters} and its variants \cite{rebuffi2017learning,rebuffi2018efficient}, {\em Deep Adaptation Networks (DAN)} \cite{rosenfeld2017incremental}, and {\em Batch Norm Adaptation (BN Adapt)} \cite{bilen2017universal}, which are explicitly designed to minimize the number of model parameters, while our method sits at the other end of the spectrum, with a focus on accuracy instead of parameter reduction. We also compare with training from scratch using Residual Adapters ({\em Scratch+}), as well as the high-capacity version of Residual Adapters described in \cite{rebuffi2017learning}, which have a similar number of parameters as SpotTune.

\noindent \textbf{Pre-trained model.} For comparing SpotTune with fine-tuning baselines in Table \ref{table:results}, we use ResNet-50 pre-trained on ImageNet, which starts with a convolutional layer followed by 16 residual blocks. The residual blocks contain three convolutional layers and are distributed into 4 segments (i.e, [3, 4, 6, 3]) with downsampling layers in between. We use the pre-trained model from Pytorch which has a classification accuracy of 75.15\% on ImageNet. For the Visual Decathlon Challenge, in order to be consistent with previous works, we adopt ResNet-26 with a total of 12 residual blocks, organized into 3 segments (i.e., [4, 4, 4]). The channel size of each segment is 64, 128, 256, respectively. We use the ResNet-26 pre-trained on ImageNet provided by \cite{rebuffi2018efficient}.

\vspace{0.05in}
\noindent \textbf{Policy network architecture.}
For the experiments with ResNet-50 (Table \ref{table:results}), we use a ResNet with 4 blocks for the policy network. The channel size of each block is 64, 128, 256, 512, respectively. For the Visual Decathlon Challenge with ResNet-26, the policy network consists of a ResNet with 3 blocks. The channel size of each block is 64, 128, 256, respectively. 

\vspace{0.05in}
\noindent \textbf{Implementations details.} Our implementation is based on Pytorch. All models are trained on 2 NVIDIA V100 GPUs. For comparing SpotTune with fine-tuning baselines, we use SGD with momentum as the optimizer. The momentum rate is set to be 0.9, the initial learning rate is 1e-2 and the batch size is 32. The initial learning rate of the policy network is 1e-4. We train the network with a total of 40 epochs and the learning rate decays twice at 15th and 30th epochs with a factor of 10.

For the Visual Decathlon Challenge, we also use SGD with momentum as the optimizer. The momentum rate is 0.9 and the initial learning rate is 0.1. The batch size is 128.  The initial learning rate of the policy network is 1e-2. We train the network with a total of 110 epochs and the learning rate decays three times at 40th, 60th and 80th epochs with a factor of 10. We freeze the first macro blocks (4 residual blocks) of the ResNet-26 and only apply the adaptive fine-tuning for the rest of the residual blocks. This choice reduces the number of parameters and has a regularization effect. The temperature of the Gumbel-Softmax distribution is set to 5 for all the experiments. Our source code will be publicly available.

%\vspace{0.2in}
\subsection{Results and Analysis}
\subsubsection{SpotTune vs. Fine-tuning Baselines}
The results of SpotTune and the fine-tuning baselines are listed in Table \ref{table:results}. Clearly, SpotTune yields consistently better results than other methods. Using the pre-trained model on ImageNet as a {\em feature extractor} (with all parameters frozen) can reduce the number of parameters when the model is applied to a new dataset, but it leads to bad performance due to the domain shift. All the fine-tuning variants ({\em Standard Fine-tuning, Stochastic Fine-tuning, Fine-tuning last-k}) achieve higher accuracy than the {\em Feature Extractor} baseline, as expected. Note that the results of {\em Fine-tuning last-k} show that manually deciding the number of layers to fine-tune may lead to worse results than {\em standard fine-tuning}. 
The {\em Fine-tuned ResNet-101} has higher capacity and thus performs better than the other fine-tuning variants. Although it has twice as many fine-tuned blocks and is significantly more computationally expensive than SpotTune, it still performs worse than our method in all datasets, except in WikiArt. We conjecture this is because WikiArt has more training examples than the other datasets. To test this hypothesis, we evaluated both models when 25\% of the WikiArt training data is used. In this setting, SpotTune achieves 61.24\% accuracy compared to 60.20\% of the fine-tuned ResNet-101. This gap increases even more when 10\% of the data is considered (49.59\% vs. 47.05\%).

By inducing the fine-tuned models to be close to the pre-trained model, $L^2$-SP achieves better results than other fine-tuning variants, but it is inferior to SpotTune in all datasets. However, it should be noted that $L^2$-SP is complementary to SpotTune and can be combined with it to further improve the results.

SpotTune is different from all the baselines in two aspects. On one hand, the fine-tuning policy in SpotTune is specialized for each instance in the target dataset. This implicitly takes the similarities between the images in the target dataset and the source dataset into account. On the other hand, sharing layers with the source task without parameter refinement reduces overfitting and promotes better re-use of features extracted from the source task. 
We also consider two variants of SpotTune in the experiments. The first one is {\em SpotTune (running fine-tuned blocks)} in which during testing all the images are routed through the fine-tuned blocks. With this setting, the accuracy drops on all the datasets. This suggests that certain images in the target data can benefit from reusing some of the layers of the pre-trained network. The second variant is {\em SpotTune (global-k)} in which we set $k$ to 3 in the experiments. Generally, SpotTune (global-3) performs worse than SpotTune, but is around 3 times more compact and, interestingly, is better than {\em Fine-tuning last-3}. This suggests that it is beneficial to have an image-specific fine-tuning strategy, and manually selecting the last $k$ layers is not as effective as choosing the optimal non-contiguous set of $k$ layers for fine-tuning.
\def\arraystretch{1.2}%  1 is the default, change whatever you need
\begin{table*}[!htb]
	\small
	\begin{center}
		\begin{tabular}{c  c c c c c c c c c c c c c c} 
\hline
Model	& \#par  & ImNet  & Airc. & C100  & DPed & DTD & GTSR & Flwr & OGlt & SVHN  &UCF   & Score \\
\hline
Scratch	& 10x  & 59.87  & 57.10 & 75.73 & 91.20 & 37.77 & 96.55  & 56.30  & 88.74 &96.63  & 43.27 & 1625   \\

Scratch+ \cite{rebuffi2017learning} & 11x  & 59.67  & 59.59 & 76.08 & 92.45 & 39.63 & 96.90  & 56.66 & 88.74 & 96.78 & 44.17 & 1826   \\

Feature Extractor& 1x  & 59.67  &  23.31 & 63.11 & 80.33  & 55.53 & 68.18 & 73.69  & 58.79  & 43.54   & 26.80 & 544 \\

Fine-tuning	\cite{rebuffi2018efficient} & 10x  & 60.32 & 61.87 & 82.12 & 92.82 & 55.53 &99.42  & 81.41  & 89.12  & 96.55 & 51.20 & 3096    \\

BN Adapt. \cite{bilen2017universal}	&1x  & 59.87  & 43.05 & 78.62 & 92.07& 51.60& 95.82 & 74.14  & 84.83  & 94.10  & 43.51 &  1353  \\

LwF	\cite{li2017learning} &  10x & 59.87 & 61.15 & 82.23 & 92.34 & 58.83 & 97.57  & 83.05  & 88.08  & 96.10  & 50.04 &2515    \\

Series Res. adapt. \cite{rebuffi2017learning}	& 2x & 60.32 & 61.87  & 81.22&  93.88& 57.13   & 99.27  & 81.67  &  89.62 & 96.57& 50.12 & 3159  \\

Parallel Res. adapt. \cite{rebuffi2018efficient} & 2x  & 60.32 & 64.21 & 81.92 & 94.73 & 58.83 & 99.38 & 84.68   & 89.21 & 96.54& 50.94 & 3412  \\

Res. adapt. (large) \cite{rebuffi2017learning} &12x & 67.00 & 67.69 & 84.69 & 94.28  & 59.41 & 97.43 & 84.86 & 89.92  & 96.59 & 52.39 & 3131 \\

Res. adapt. decay \cite{rebuffi2017learning} & 2x & 59.67 & 61.87 & 81.20  & 93.88 & 57.13 & 97.57& 81.67 & 89.62   & 96.13 & 50.12 & 2621  \\
Res. adapt. finetune all \cite{rebuffi2017learning} & 2x &  59.23 & 63.73  & 81.31  & 93.30 & 57.02  & 97.47  &  83.43   & 89.82&  96.17 &  50.28 & 2643 \\

DAN	\cite{rosenfeld2017incremental} & 2x  & 57.74 & 64.12  & 80.07& 91.30 & 56.54 & 98.46  & 86.05  & 89.67  & 96.77  & 49.48 & 2851   \\

PiggyBack \cite{mallya2018piggyback}	&  1.28x & 57.69  & 65.29 & 79.87 & 96.99  & 57.45 & 97.27  & 79.09  & 87.63  & 97.24   & 47.48  & 2838 
\\
\hline
SpotTune (Global-k) & 4x & 60.32 & 61.57  & 80.30 & 95.78  & 55.80 &  99.48 & 85.38  &  88.41 & 96.47  & 51.05 & 3401 \\

SpotTune & 11x & 60.32  & 63.91 & 80.48 &  96.49  & 57.13  & 99.52 &  85.22 & 88.84  &  96.72  & 52.34   & \textbf{3612}  \\
\hline
			
\end{tabular}
\end{center}
\caption{Results of SpotTune and baselines on the Visual Decathlon Challenge. The number of parameters is specified with respect to a ResNet-26 model as in \cite{rebuffi2017learning}.}
\label{table:results_vis}
\end{table*}

\vspace{-0.15in}
\subsubsection{Visualization of Policies}
\begin{figure}[!htb]
    \centering
\includegraphics[width=0.43\textwidth]{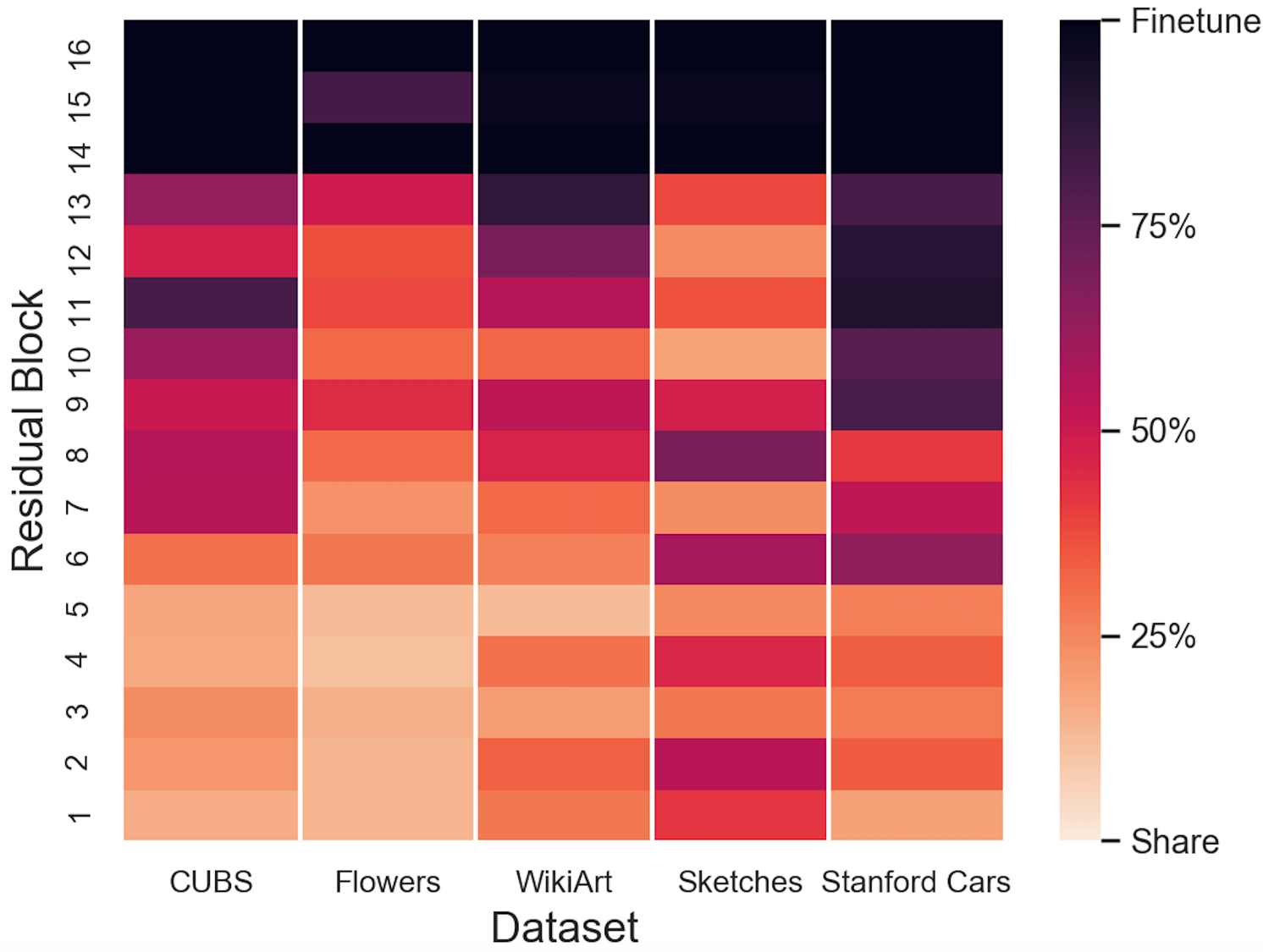}    
    \caption{Visualization of policies on CUBS, Flowers, WikiArt, Sketches and Stanford Cars. Note that different datasets have very different policies. SpotTune automatically identifies the right fine-tuning policy for each dataset, for each training example.  }
    \label{fig:policies}
\end{figure}

To better understand the fine-tuning policies learned by the policy network, we visualize them on CUBS, Flowers, WikiArt, Sketches, and Stanford Cars in Figure \ref{fig:policies}. The polices are learned on a ResNet-50 which has 16 blocks. The tone of red of a block indicates the number of images that were routed through the fine-tuned path of that block. For example, a block with a dark tone of red and a 75\% level of fine-tuning (as shown in the scale depicted in the right of Figure \ref{fig:policies}) means 75\% of the images in the test set use the fine-tuned block and the remaining 25\% images share the pre-trained ImageNet block. The illustration shows that different datasets have very different fine-tuning policies. SpotTune allows us to automatically identify the right policy for each dataset, as well as for each training example, which would be infeasible through a manual approach.

%\vspace{-0.3cm}
\subsubsection{Visualization of Block Usage}
\begin{figure}[!htb]
    \centering
\includegraphics[width=0.5\textwidth]{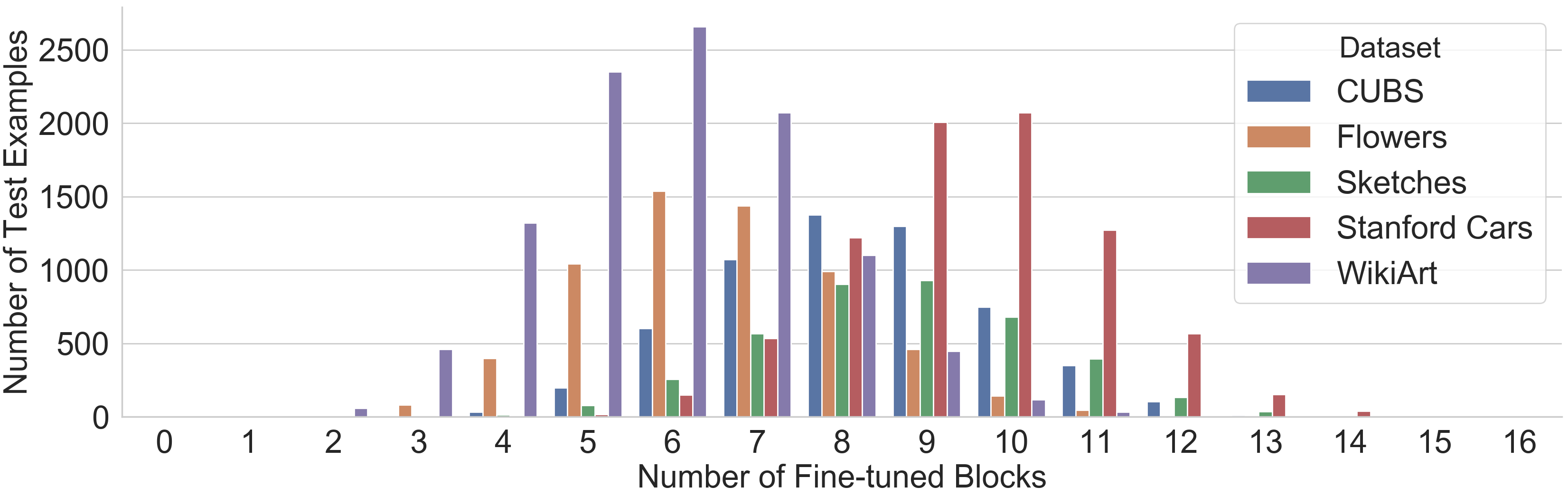}    
    \caption{Distribution of the number of fine-tuned blocks used by the test examples in the datasets. Different tasks and images require substantially different fine-tuning for best results, and this can be automatically inferred by SpotTune.%For each dataset, different images tend to use a different number of fine-tuned blocks, and the usage distribution differs across datasets. 
    }
    \label{fig:blocks}
\end{figure}

Besides the learned policies for each residual block, we are also interested in the number of fine-tuned blocks used by each dataset during testing. This can reveal the difference of the distribution of each target dataset and can also shed light on how the policy network works. In Figure \ref{fig:blocks}, we show the distribution of the number of fine-tuned blocks used by each target dataset. During testing, for each dataset we categorize the test examples based on the number of fine-tuned blocks they use. For example, from Figure \ref{fig:blocks}, we can see around 1000 images in the test set of the CUBS dataset use 7 fine-tuned blocks. 

We have the following two observations based on the results. First, for a specific dataset, different images tend to use a different number of fine-tuned blocks. This again validates our hypothesis that it is more accurate to have an image-specific fine-tuning policy rather than a global fine-tuning policy for all images. Second, the distribution of fine-tuned blocks usage differs significantly across different target datasets. This demonstrates that based on the characteristics of the target dataset, standard fine-tuning (which optimizes all the parameters of the pre-trained network towards the target task) may not be the ideal choice when conducting transfer learning with convolutional networks. 

\begin{figure}[!htb]
    \centering
\includegraphics[width=0.48\textwidth]{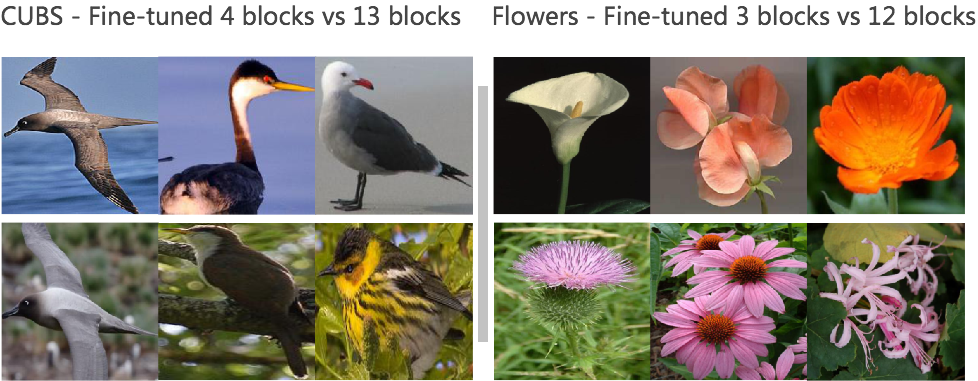}    
    \caption{Example images that use a small and large number of fine-tuned blocks. %The first row shows sample images that fine-tune 4 blocks on CUBS and 3 blocks on Flowers, respectively. The second row shows sample images that fine-tune 13 blocks on CUBS and 12 blocks on Flowers, respectively. 
    }
    \label{fig:images}
\end{figure}

Figure \ref{fig:images} shows example images that use a different number of fine-tuned blocks on CUBS and Flowers. We observe that images that use a small number of fine-tuned blocks tend to have a cleaner background (possibly due to similarity with ImageNet data), while images that use a large number of fine-tuned blocks often have a more complex background. An interesting area for future work is to quantify the interpretability of both pre-trained and fine-tuned convolutional filters using e.g., \emph{Network Dissection}  \cite{bau2017network}, in order to better understand these visual patterns.
%We consider adopting  to quantify the evolution of representations under SpotTune and leave this as future work.
%\vspace{-0.3cm}
\vspace{-0.4cm}

%\vspace{-0.38cm}
\subsubsection{Visual Decathlon Challenge}
 We show the results of SpotTune and baselines on the Visual Decathlon Challenge in Table \ref{table:results_vis}. Among all the baselines, SpotTune achieves the highest Visual Decathlon score. 
 Compared to standard fine-tuning, SpotTune has almost the same amount of parameters and improves the score by a large margin (3612 vs 3096). Considering the Visual Decathlon datasets, and the 5 datasets from our previous experiments, SpotTune shows superior performance on 12 out of 14 datasets over standard fine-tuning. 
 Compared with other recently proposed methods on the Visual Decathlon Challenge \cite{mallya2018piggyback,rosenfeld2017incremental,rebuffi2017learning,rebuffi2018efficient,li2017learning}, SpotTune sets the new state of the art for the challenge by only exploiting the transferability of the features extracted from ImageNet, without changing the network architecture. This is achieved without bells and whistles, i.e., we believe the results could be even further improved with more careful parameter tuning, and the use of other techniques such as data augmentation, including jittering images at test time and averaging their predictions.
 
 In SpotTune (Global-k), we fine-tune 3 blocks of the pre-trained model for each task which greatly reduces the number of parameters and still preserves a very competitive score.  Although we focus on accuracy instead of parameter reduction in our work, we note that training our global-k variant with a multi-task loss on all 10 datasets, as well as model compression techniques, could further reduce the number of parameters in our method. We leave this research thread for future work.

%------------------------------------------------------------------------
\section{Conclusion}
We proposed an adaptive fine-tuning algorithm called SpotTune which specializes the fine-tuning strategy for each training example of the target dataset.  We showed that our method outperforms the key most popular and widely used protocols for fine-tuning on a variety of public benchmarks. We also evaluated SpotTune on the Visual Decathlon challenge, achieving the new state of the art, as measured by the overall score across the 10 datasets. 
%We see many interesting directions as follow up work, including extending our method to learn feature sharing across tasks, dealing with layerwise selection of multiple pre-trained models, and joint source and target domain modeling when data from the source domain is available.
%------------------------------------------------------------------------
%%%%%%%%% REFERENCE 
\\
\\
\noindent{ \textbf{Acknowledgements.}}
We would like to thank Professor Song Han for helpful discussions. This work is in part supported by the Intelligence Advanced Research Projects Activity (IARPA) via Department of Interior/ Interior Business Center (DOI/IBC) contract number D17PC00341. This work is also in part supported  by CRISP, one of six centers in JUMP, an SRC program sponsored by DARPA, and NSF CHASE-CI \#1730158. The U.S. Government is authorized to reproduce and distribute reprints for Governmental purposes notwithstanding any copyright annotation thereon. Disclaimer: The views and conclusions contained herein are those of the authors and should not be interpreted as necessarily representing the official policies or endorsements, either expressed or implied, of IARPA, DOI/IBC, or the U.S. Government.
{\small
\bibliographystyle{ieee}
\bibliography{egbib}
}
\end{document}